\documentclass[runningheads]{llncs}
\usepackage{multirow}
\usepackage{graphicx}
\usepackage{mathrsfs}
\usepackage{subfigure} 
\usepackage{marvosym}
\begin{document}
\title{SAR: Scale-Aware Restoration Learning for \\3D Tumor Segmentation}
%
%
\author{Xiaoman Zhang\inst{1,2}, Shixiang Feng\inst{1}, Yuhang Zhou\inst{1}, Ya Zhang\inst{1,2}\textsuperscript{(\Letter)}, \\ and Yanfeng Wang\inst{1}}

\authorrunning{X. Zhang et al.}

\institute{Cooperative Medianet Innovation Center, Shanghai Jiao Tong University \and
Shanghai AI Laboratory\\
{\scriptsize \email{\{xm99sjtu, fengshixiang, zhouyuhang, ya\_zhang, wangyanfeng\}@sjtu.edu.cn} 
}}

\maketitle              
\begin{abstract}
Automatic and accurate tumor segmentation on medical images is in high demand to assist physicians with diagnosis and treatment. However, it is difficult to obtain massive amounts of annotated training data required by the deep-learning models as the manual delineation process is often tedious and expertise required. Although self-supervised learning (SSL) scheme has been widely adopted to address this problem, most SSL methods focus only on global structure information, ignoring the key distinguishing features of tumor regions: local intensity variation and large size distribution. In this paper, we propose Scale-Aware Restoration (SAR), a SSL method for 3D tumor segmentation. Specifically, a novel proxy task, i.e. scale discrimination, is formulated to pre-train the 3D neural network combined with the self-restoration task. Thus, the pre-trained model learns multi-level local representations through multi-scale inputs. Moreover, an adversarial learning module is further introduced to learn modality invariant representations from multiple unlabeled source datasets. We demonstrate the effectiveness of our methods on two downstream tasks: i) Brain tumor segmentation, ii) Pancreas tumor segmentation. Compared with the state-of-the-art 3D SSL methods, our proposed approach can significantly improve the segmentation accuracy. Besides, we analyze its advantages from multiple perspectives such as data efficiency, performance, and convergence speed.
\end{abstract}

\section{Introduction}
Automatic and accurate tumor segmentation on medical images is an essential step in computer-assisted clinical interventions.
Deep learning approaches have recently made remarkable progress in medical image analysis \cite{Litjens2017ASO}. However, the success of these data-driven approaches generally demands massive amounts of annotated training data \cite{Shin2016}. Unfortunately, obtaining a comprehensively annotated dataset in the medical field is extremely challenging due to the considerable labor and time costs as well as the expertise required for annotation. 

To reduce the requirement on labeled training data, self-supervised learning (SSL) scheme has been widely adopted for visual representation learning \cite{Chen2019SelfSupervisedLF,DoerschGE15,Spyros07728}. A generic framework is to first employ predefined self-supervision tasks to learn feature representations capturing the data characteristics and further fine-tune the model with a handful of task-specific annotated data. For medical image analysis, many existing 3D SSL methods borrow ideas from proxy tasks proposed for natural images and extend them to 3D vision, such as rotating \cite{Chen2019SelfSupervisedLF}, jigsaw puzzles \cite{Zhuang2019SelfSupervisedFL} and contrastive learning \cite{Taleb2020}. Another stream of 3D SSL methods \cite{Haghighi2020LearningSR,Parts2Whole} follows Model Genesis \cite{Zongwei2019}, one of the most famous SSL methods for medical images, which pre-trains the network with a transform-restoration process. 

Despite their promises in medical image analysis, existing 3D SSL methods mainly have two drawbacks when applying to tumor segmentation. Firstly, different from ordinary semantic segmentation, tumor segmentation  focuses more on the local intensity and texture variations rather than global structural or semantic information.
Representations learned by SSL methods extended directly from computer vision tasks may not be effectively transferred to this target task.

\begin{figure}[tb]
\includegraphics[width=\textwidth]{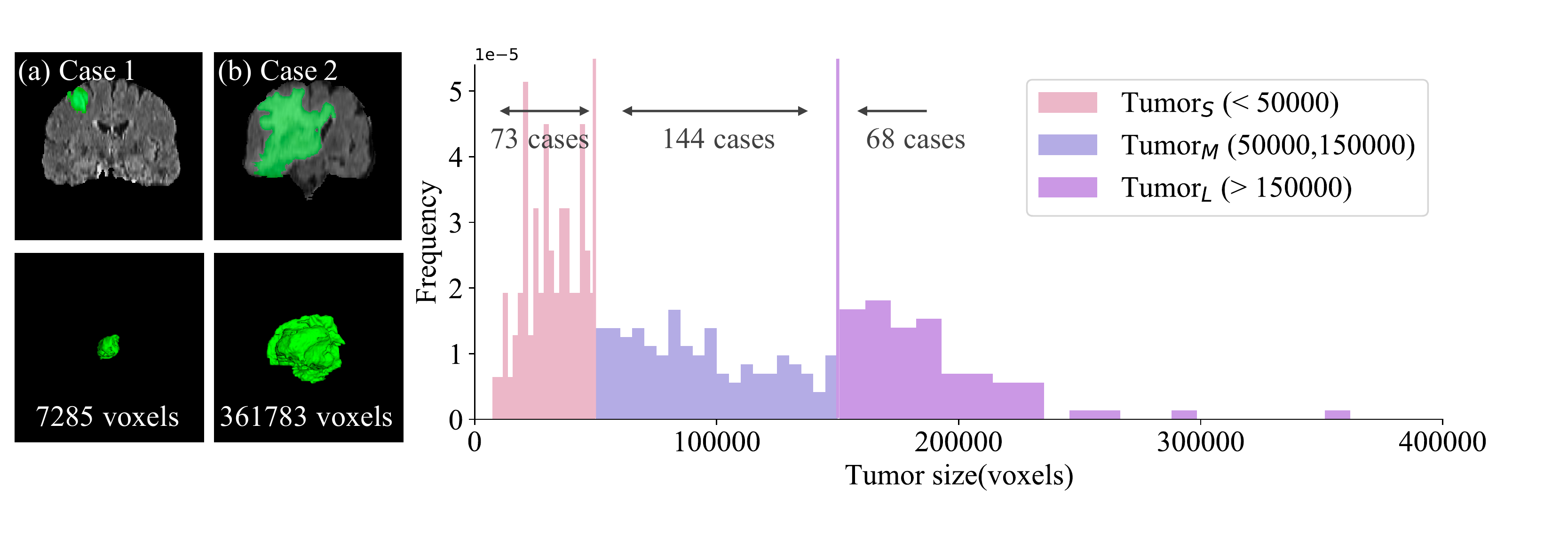}
\caption{Examples of brain MRI scans showing the large variations of shape and size of the tumor (left). 
Tumor size distribution for the BraTS2018 dataset (right).}
\label{fig1}
\end{figure}
Secondly, the tumor region in CT/MRI scans, as shown in Fig.~\ref{fig1}, can vary in scale, appearance and geometric properties. In particular, the largest tumor in the dataset occupies more than hundreds of thousands of voxels, but the smallest one has even less than one thousand voxels. 
Such a large range of tumor scales increases the difficulty of training a segmentation model.
Thus, learning to capture visual patterns of various scales is expected to beneficial for tumor segmentation, which is not taken into account by most existing methods. 

In this paper, we propose Scale-Aware Restoration (SAR) learning, a self-supervised method based on the prior-knowledge for 3D tumor segmentation.
Specifically, we propose a novel proxy task, i.e. scale discrimination, to pre-train 3D neural networks combined with the self-restoration task. Thus, the pre-trained model is encouraged to learn multi-scale visual representations and low-level detailed features.
Moreover, since multiple unlabeled source datasets are used in the pre-training stage, we further introduce an adversarial learning module to learn modality-invariant representations.

\section{Method}
\begin{figure}[tb]
\includegraphics[width=\textwidth]{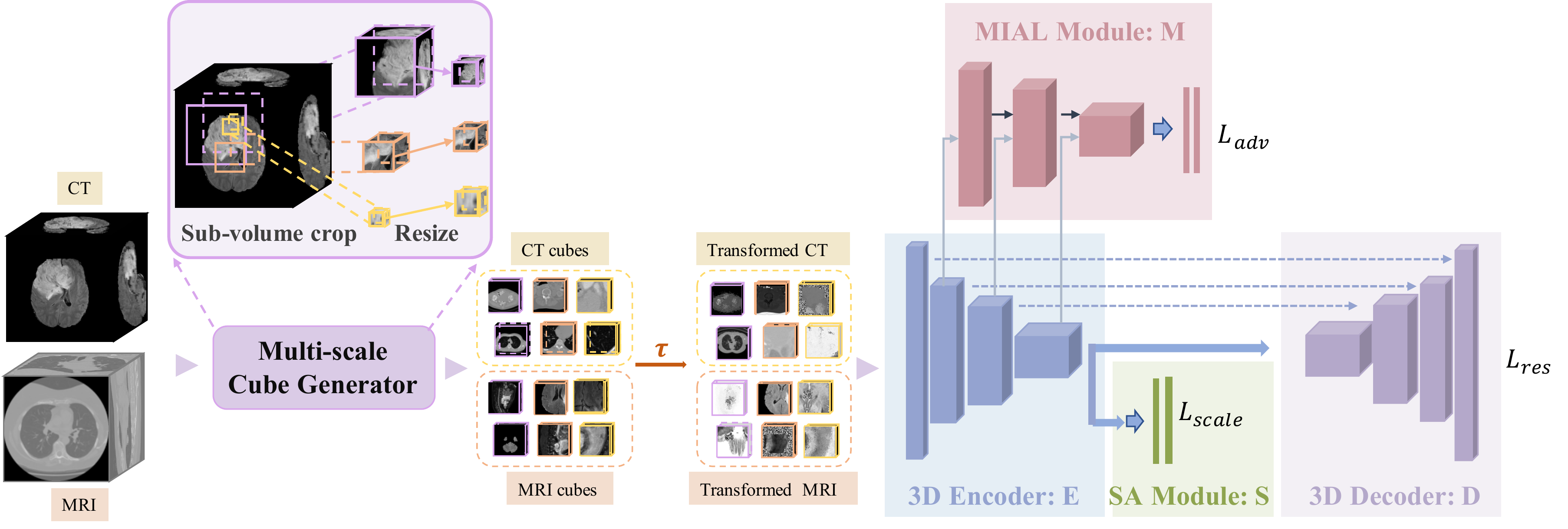}
\caption{Schematic diagram of the proposed method.}
\label{fig2}
\end{figure}

In this section, we introduce the proposed Scale-Aware Restoration (SAR) learning method in detail.
As shown in Fig.~\ref{fig2}, given an unlabeled image $x$, we extract multi-scale sub-volumes, denoted by $x_i$, which are then fed into the transformation module to get the transformed inputs $\hat{x}_i$. The goal of the proposed self-supervised proxy task, SAR, is to learn to predict the scale information (Scale-Aware) and recover the origin sub-volume (Restoration) of the transformed inputs $\hat{x}_i$. Moreover, we design an adversarial learning module to capture modality-invariant features as the pre-trained source datasets have various modalities and naively fusing multi-modal representations may lead to transfer performance degradation. 

\subsection{Self-Restoration Framework}
The goal of the self-restoration task is to learn to recover origin sub-volume $x_i$ from transformed input $\hat{x_i}$, where $\hat{x_i}=\tau(x_i)$ and $\tau(\cdot)$ denotes the transformation function. In practice, we apply a sequence of image transformations to the input sub-volume. We follow \cite{Zongwei2019} and adopt the same set of transformations, i.e. non-linear, local-shuffling, inner and outer-painting, corresponding to the appearance, texture and context, respectively. We take 3D U-Net, the most commonly used network structure for 3D medical image analysis, as the backbone.
The Mean Square Error (MSE) loss is employed as the optimizer function for the restoration task, which is expressed as follows:
\begin{equation} 
\min \limits_{\theta_E,\theta_D} L_{res}= \sum_{{x_i}\in\mathcal{D}} {\left\| D(E(\hat{x_i})) - x_i \right\|}_{2},
\label{loss_mse} 
\end{equation}
where $\theta_E$ and $\theta_D$ are parameters of the encoder $E$ and the decoder $D$, respectively.

\subsection{Scale Aware Module}
The presence of tumors of various sizes increases the difficulty of training a segmentation model, which motivates us to pre-train a model that can capture multi-scale visual patterns. 
Accordingly, we propose a novel self-supervised learning task, scale discrimination, in which the model learns to predict the scale information of multi-scale sub-volumes.
The multi-scale cube generator in Fig.~\ref{fig2} shows the pre-processing process for input volumes. For each volume in the training dataset $\mathcal{D}$, we first randomly crop sub-volumes of three different scales: large, medium and small, corresponding to approximately ${1/2}, 1/4$ and $1/8$ of the side length of the original cube respectively. 
Since the smallest tumor occupies nearly 1/8 of the whole volume while the largest has nearly 1/2, we pick these scales to make sure the learnt representation is aware of the texture and intensity over the range of tumor size.
The numbers of large scale, medium scale and small scale are in the ratio $1:1:2$ as the small scale cubes can be more diverse. The exact number of cubes are shown in Table~\ref{table_dataset}. 
To fit the input size of the model, we then resize all the sub-volumes to the same size: $64\times64\times32$. 

Let $x_i$ denotes the resized input sub-volume and $y_i$ is the corresponding scale label. We build a Scale-Aware (SA) module $S$ to predict the scale of input sub-volumes, ensuring the learned representations containing multi-level discriminative features. Speciﬁcally, features generated from Encoder $E$ are input to the SA module $S$ for scale prediction. The optimizer function for the SA module is defined as follows:
\begin{equation} 
\min \limits_{\theta_E,\theta_S}L_{scale}=\sum_{(x_i,y_i)\in\mathcal{D}}L_{CE}(S(E(\hat{x_i})),y_i),
\label{eq2} 
\end{equation}
where $L_{CE}(\cdot)$ represents cross entropy loss, $\hat{x_i}$ is the transformed input volume, $y_i$ is the corresponding scale label, and $\theta_S$ are parameters of the SA module $S$.

\subsection{Modality Invariant Adversarial Learning}
Medical images in different modalities are quite different in characteristics. To avoid the performance degradation caused by cross-modality generalization, we develop a Modality Invariant Adversarial Learning (MIAL) module $M$ for modality invariant representation learning. 
Different from the common domain discriminator, which uses $E(\hat{x})$ as input, we contact features of different layers in $E$ with the corresponding layer in $M$. 
Such a densely connected operation improves the modality invariant feature learning by strengthening the connection between the encoder and discriminator. 
If the discriminator can successfully classify the modal of input sub-volumes, it means that the features extracted from the encoder still contain modality characteristics. Hence, we apply the min-max two-player game, where the MIAL module $M$ is trained to distinguish the modality of input sub-volumes, and the encoder $E$ is trained simultaneously to confuse the discriminator $M$. 

Denote $\mathcal{D}=\{\mathcal{D}_{CT}, \mathcal{D}_{MRI}\}$ as the training dataset. The following adversarial loss is applied:
\begin{equation}
	\min\limits_{\theta_E}\max\limits_{\theta_M}L_{adv}=\sum_{x_{i}^{CT}\in \mathcal{D}_{CT}} log (M(E(\hat{x}_{i}^{CT})))+\\
	\sum_{x_{i}^{MRI}\in \mathcal{D}_{{MRI}}} log(1- M(E(\hat{x}_{i}^{MRI}))),
\end{equation}
where ${\theta_E}$ and ${\theta_M}$ are parameters of encoder $E$ and MIAL module $M$, $\hat{x}_{i}^{CT}$ and $\hat{x}_{i}^{MRI}$ are the transformed input sub-volume of corresponding modality.

\subsection{Objective Function}
The final objective function for optimizing the pre-trained model can thus be formulated as follows:
\begin{equation}
	\min\limits_{\theta_E,\theta_D,\theta_S}\max\limits_{\theta_M}L_{adv}+\alpha L_{scale}+ \beta L_{res},
\end{equation}
where $\alpha$ and $\beta$ are trade-off parameters weighting the loss functions, since the magnitudes of the three loss functions are inconsistent. In our experiments, we empirically set $\alpha=1,\beta=10$ to keep the order of magnitudes consistently. 

\section{Experiments}
\subsection{Pre-training Datasets}
We collected 734 CT scans and 790 MRI scans from 4 public datasets, including LUNA2016 \cite{Setio2015AutomaticDO}, LiTS2017 \cite{bilic2019liver}, BraTS2018 \cite{Bakas2018}, and MSD (heart) \cite{simpson2019large}, as source datasets to build the pre-trained model, SAR.
These datasets are quite diverse, containing two modalities (i.e. MRI and CT) of data, from three different scan regions (i.e. brain, chest, and abdomen), and targeting four organs (i.e. lung, liver, brain, and heart). The detailed information about the source datasets is summarized in Table~\ref{table_dataset}. 

\begin{table}[!hbt]
\centering
\caption{Statistics of the pre-training datasets where cases represent volumes used in the pre-trained process and sub-volume number represents the number of cropped cubes at different scales.}
\setlength{\tabcolsep}{3pt}
\begin{tabular}{c|c|c|c|c|c}
\hline
Dataset& Cases & Modality & Organ & Median Shape &  Sub-volume Number \\
\hline
LUNA2016 & 623 & CT & Lung & $238\times 512 \times 512 $ &  (19936, 9968, 9968)\\
LiTS2017 & 111 & CT & Liver & $432 \times 512 \times 512$ & (7104, 3552, 3552) \\
BraTS2018 & 190x4 & MRI & Brain &$138 \times 169 \times 138$ & (24320, 12160, 12160) \\
MSD (heart) & 30 & MRI & Heart & $115 \times 320 \times 232$ &  (1920, 960, 960) \\
\hline
\end{tabular}
\label{table_dataset}
\end{table}

To alleviate the heterogeneity caused by multi-center (e.g. due to different scanners and imaging protocols), we process the data with spatial and intensity normalization. We resample all volumes to the same voxel spacing ($1\times 1 \times 1 mm^3$) by using third-order spline interpolation. To mitigate the side-effect of extreme pixel value, we clip the intensity values of CT volumes on the min (-1000) and max (+1000) interesting Hounsfield Unit range. And for the MRI volumes, all the intensity values are clipped on the min (0) and max (+4000) interesting range. Each volume is normalized independently to [0,1].

\subsection{Downstream Tasks}
\subsubsection{Brain tumor segmentation}
We perform the brain tumor segmentation task using the training set of BraTS2018 \cite{Bakas2018}. The dataset contains 285 cases acquired with different MRI scanners. 
Following the experiment strategy in \cite{Zongwei2019}, we use FLAIR images and only segment the whole brain tumor. We randomly divide the training and test sets in the ratio of 2:1. Note that all the test images for target tasks are not exposed to model pre-training. For data preprocessing, each image is independently normalized by subtracting the mean of its non-zero area and dividing by the standard deviation. We report test performance through the mean and standard deviation of five trials. 

\subsubsection{Pancreas organ and tumor segmentation}
We perform the pancreas tumor segmentation task using the Pancreas dataset of MSD \cite{simpson2019large}, which contains 282 3D CT training scans. Each scan in this dataset contains 3 different classes: pancreas (class 1), tumor (class 2), and background (class 0). For data preprocessing \cite{isensee2019automated}, we clip the HU values in each scan to the range $[-96, 215]$, and then subtract 77.99 and finally divide by 75.40. We report the mean and standard deviation of five-fold cross-validation segmentation results. 

\subsection{Implementation and Training Details}
In the pre-training stage, the basic encoder and decoder follow the 3D U-Net \cite{iek20163DUL} architecture with batch normalization. The MIAL module is constructed with three convolution layers with a global average pooling (GAP) layer and two fully connected (FC) layers. The SA module simply consists of a GAP layer and two FC layers. The network is trained using SGD optimizer, with an initial learning rate of 1e0, 1e-1, 1e-3 for 3D U-Net, SA module and MIAL module respectively. We use ReduceLROnPlateau to schedule the learning rate according to the validation loss.

In the fine-tuning stage, we transfer the 3D encoder-decoder for the target segmentation tasks with dice loss. We use the Adam optimizer \cite{Kingma2014} with an initial learning rate of 1e-3 for fine-tuning and use ReduceLROnPlateau to schedule the learning rate. Note that all of the layers in the model are trainable during fine-tuning. All experiments were implemented with a PyTorch framework on GTX 1080Ti GPU. 

\section{Results}
\subsubsection{Comparison with state-of-the-art methods}
In this section, we compare SAR with training from scratch and the state-of-the-art 3D SSL methods including Model Genesis (Genesis) \cite{Zhou2020ModelsG}, Relative 3D patch location (3D-RPL), 3D Jigsaw puzzles (3D-Jig), 3D Rotation Prediction (3D-Rot) and 3D Contrastive Predictive Coding (3D-CPC) \cite{Taleb2020}. For a fair comparison, we re-trained all the SSL methods under the same experiment setup, including the network architecture and the pre-training datasets. For SSL tasks that use only the encoder during pre-training, we randomly initialize the decoder during fine-tuning. 

\begin{table}[ht]
\centering
\caption{Comparison with state-of-the-art methods.}
\setlength{\tabcolsep}{3pt}
\begin{tabular}{l|ccc|c|cc}
\hline
& \multicolumn{4}{c|}{ Task 1.BraTS2018 } &  \multicolumn{2}{c}{ Task 2.MSD(Pancreas)} \\
\hline
& Tumor$_{S}$ & Tumor$_{M}$ & Tumor$_{L}$ & Tumor$_{all}$ & Organ & Tumor\\
\hline
Scratch &$65.41_{\pm 7.82}$ & $78.30_{\pm3.33}$ & $74.90_{\pm4.72}$ & $74.35_{\pm 3.95}$ & $73.42_{\pm 1.21 }$ & $25.19_{\pm 4.27 }$ \\
\hline
Genesis & $74.36_{\pm 6.72}$ & $85.67_{\pm 2.04}$ & $85.62_{\pm 1.27}  $ & $83.03_{\pm 3.14}$ & $74.17_{\pm 2.38}$ & $33.12_{\pm3.08}$ \\
3D-RPL  & $72.77_{\pm 1.59}$ & $82.22_{\pm 0.29} $& $85.62_{\pm 1.27}$ & $80.15_{\pm 0.47}$ &  $71.84_{\pm 2.75 }$ & $28.30 _{\pm 3.47}$ \\
3D-Jig  & $61.24_{\pm 3.32}$ & $76.05_{\pm 1.93} $ & $72.21_{\pm 1.60}$& $71.48_{\pm 1.68} $ & $73.65_{\pm 1.58 }$ & $28.82_{\pm3.57  }$ \\
3D-Rot  & $63.06_{\pm 3.08} $ & $77.18_{\pm 1.06}$ & $72.65_{\pm 1.36} $& $72.66_{\pm 1.33} $& $70.99_{\pm 1.31 } $& $24.91_{\pm 0.63}$ \\
3D-CPC  & $65.72_{\pm 5.17} $& $77.35_{\pm 1.80}$& $72.65_{\pm 3.05}$& $73.39_{\pm 2.31} $ & $70.54_{\pm 1.39}$ & $27.99_{\pm 6.21 }$\\
\hline
+ MIAL & $76.31_{\pm 1.30}$ & $85.99_{\pm 0.68} $ & $85.20_{\pm 0.83}$ & $83.46_{\pm 0.92}$ & $75.18 _{\pm1.78}$ & $32.12_{\pm2.30}$\\
+ SA & $75.44_{\pm 3.09}$ &\textbf{87.09}$_{\pm 0.65}$& $85.74_{\pm 1.07}$& $83.95_{\pm 1.09}$ & $75.51_{\pm 0.75}$ & $32.60 _{\pm 3.02}$\\
\hline
\textbf{SAR}& \textbf{78.10}$_{\pm 3.20}$& $86.75_{\pm 1.05}$& \textbf{87.84}$_{\pm 2.45}$& \textbf{84.92}$_{\pm 1.62}$ &\textbf{75.68}$_{\pm0.85}$ & \textbf{33.92}$_{\pm3.00}$\\
\hline
\end{tabular}
\label{table_compare1}
\end{table}

\begin{figure}[bt]
\centering{\includegraphics[width=0.4\textwidth]{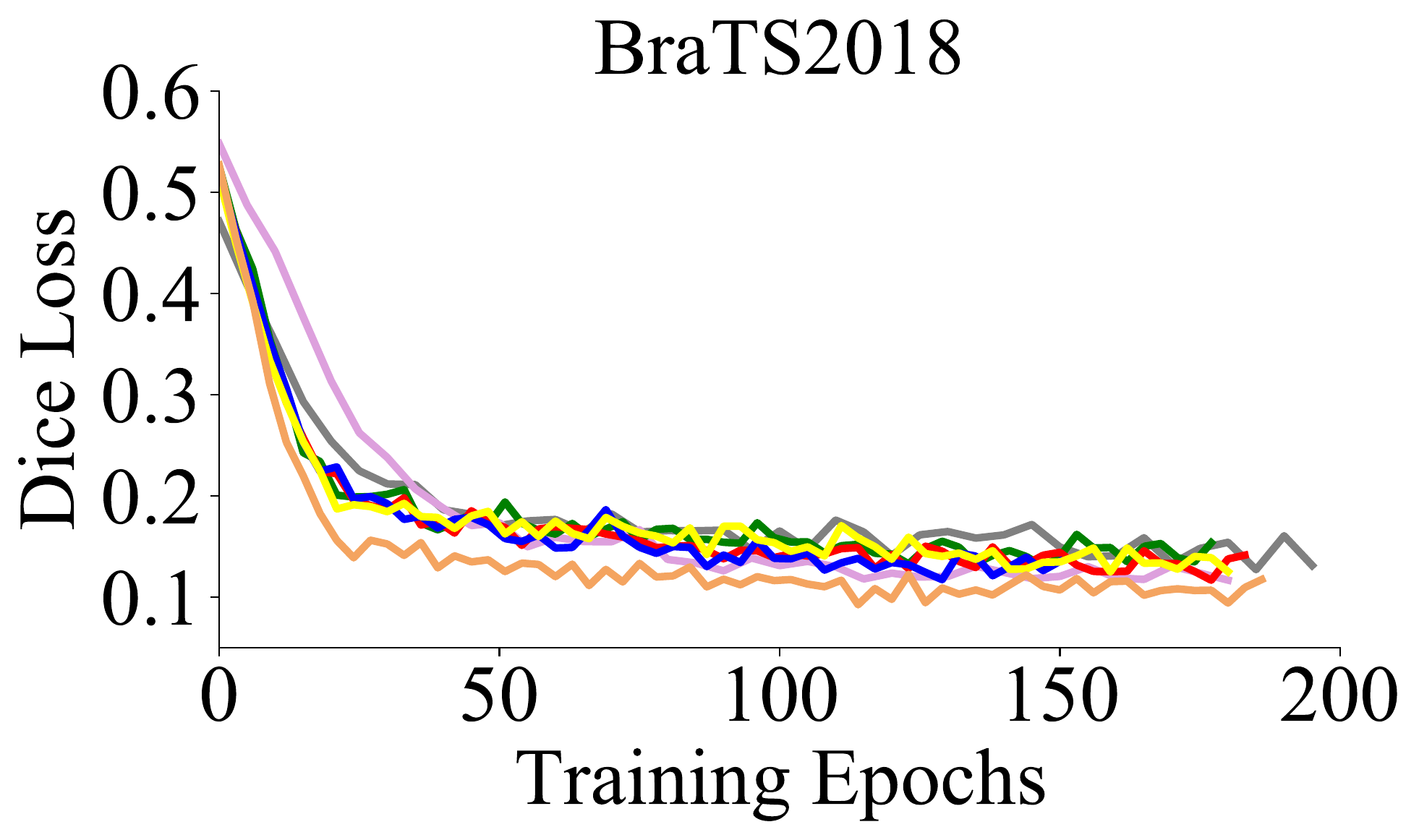}}
\centering{\includegraphics[width=0.5\textwidth]{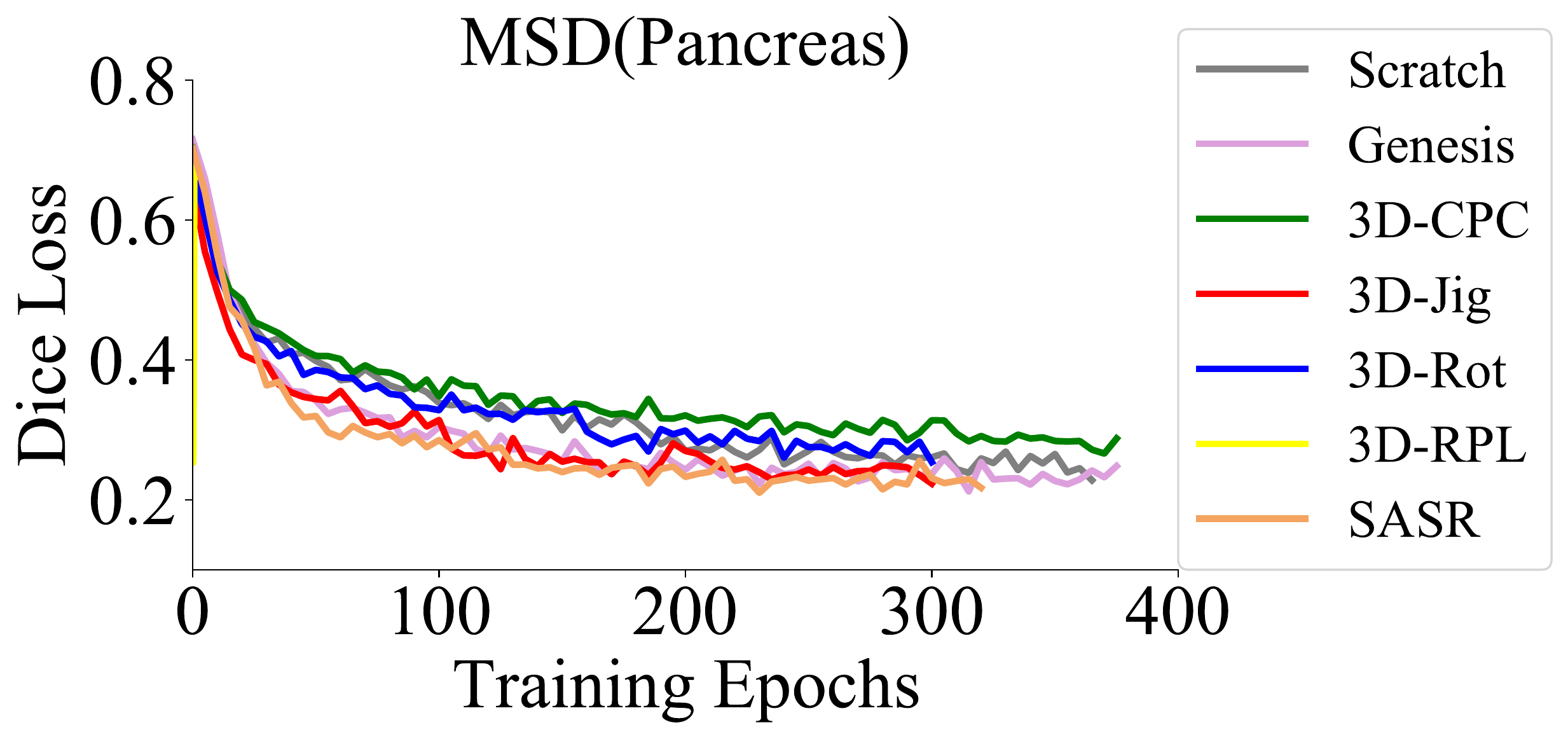}}
\caption{Comparison of the training loss curves for target tasks.} 
\label{fig_loss}
\end{figure}

The overall results are summarized in Table~\ref{table_compare1}. 
For BraTS2018, we divide the tumors into three scales according to their sizes so that we can better compare the results from a multi-scale perspective. 
Our proposed method outperforms several 3D SSL methods by a great margin and achieves a excellent performance of $84.92\%$ in average Dice for BraTS2018. 
For pancreas and tumor segmentation of MSD(Pancreas), we achieve an average Dice of $75.68\%$ for the pancreas and $33.92\%$ for the tumor, superior to other methods. Besides, we perform independent two sample t-test between the SAR vs. others. All the comparison show statistically significant results (p = 0.05) except for MSD tumor (Genesis vs. SAR)
Fig. \ref{fig_loss} shows the training loss curves compared with other methods. SAR converges faster to a stable loss and achieves a lower dice loss.
This suggests that the proposed representation learning scheme can speed up the convergence and boost performance.

\subsubsection{Ablation study}
Firstly, we conduct an ablation experiment to evaluate the effectiveness of each component proposed in SAR. In Table~\ref{table_compare1}, the performance is improved to $83.46\%$, $83.95\%$ for BraTS2018 equipped with our proposed MIAL module and SA module respectively. While combined with SA module, multi-scale information was integrated into our pre-trained model to learn comprehensive contextual and hierarchically geometric features at different levels. A signiﬁcant performance gains up to $12.69\%$ in Dice for small-size tumors in BraTS2018 conﬁrms the eﬀect of using multi-scale feature learning. 
For MSD(Pancreas), the further improvement to $75.68\%$ and $33.92\%$ of pancreas and tumor in SAR conﬁrms the adoption of the two modules yields much better results than simply combined multiple datasets.

Secondly, we validate the benefits of multi-scale representation learning on BraTS2018 in Table~\ref{table_abl1} by conducting experiments using only single-scale sub-volumes for pre-training. When we use large sub-volumes ($1/2$) to pre-train, only an average Dice of $76.17\%$ is achieved for small-scale tumors (Tumor$_{S}$), while the average Dice of large scale tumors (Tumor$_{L}$) reaches $87.88\%$. A similar situation occurs when we pre-training on the small-scale sub-volumes.
Unsurprisingly, the highest overall results are obtained by applying multi-scale sub-volumes, proving the effectiveness of learning multi-scale informative features.

\begin{table}[bt]
\centering
\caption{Ablation study results of pre-trained model using different scales of data.}
\setlength{\tabcolsep}{3pt}
\begin{tabular}{l|ccc|c}
\hline
& \multicolumn{4}{c}{BraTS2018} \\
\hline
& Tumor$_{S}$ & Tumor$_{M}$ & Tumor$_{L}$ & Tumor$_{all}$ \\
\hline
$1/2$ & 76.17$_{\pm 2.55}$  & 86.55$_{\pm0.41 }$ & \textbf{87.88 }$_{\pm 2.43}$& 84.36$_{\pm 1.69}$ \\
$1/4$ & 75.31$_{\pm 4.49} $ & 85.97$_{\pm 1.75 }$ & 86.16$_{\pm 4.67}$ & 83.44$_{\pm 3.12}$ \\
$1/8$ & 76.88$_{\pm 4.22}$ & 86.29$_{\pm 1.40}$ & 87.06$_{\pm 3.74}$ & 84.20$_{\pm 2.64}$ \\
\hline
\textbf{SAR}&\textbf{78.10}$_{\pm 3.20}$& \textbf{86.75}$_{\pm 1.05}$& 87.84$_{\pm 2.45}$& \textbf{84.92}$_{\pm 1.62}$\\
\hline
\end{tabular}
\label{table_abl1}
\end{table}

\begin{table}[bt]
\centering
\caption{Comparison of experiment results for target tasks at different ratios of annotated data.}
\setlength{\tabcolsep}{3pt}
\begin{tabular}{l|ccccc|ccccc}
\hline
& \multicolumn{5}{c|}{BraTS2018(Tumor$_{all}$)} & \multicolumn{5}{c}{MSD(Average)} \\
\hline
Proportion & $1/2$ & $1/5$ & $1/10$& $1/20$ & $1/50$ & $1/2$ & $1/5$ & $1/10$& $1/20$ & $1/50$\\
\hline
Scratch &68.37 & 57.48 & 53.57 & 41.10 & 25.36 & 41.09 & 33.24 & 30.29& 25.37 & 8.91\\ 
\textbf{SAR}&79.48 & 69.18  & 67.13 & 44.86 & 31.95 & 47.76 & 42.26 & 34.87 &  29.85 & 16.82\\
\hline
\end{tabular}
\label{table_anno}
\end{table}

\subsubsection{Analysis of annotation cost}
To validate the gains in data-efficiency, we train the model with different proportions ($\frac{1}{2}$, $\frac{1}{5}$, $\frac{1}{10}$, $\frac{1}{20}$, $\frac{1}{50}$) of the training data and use the same test set to measure the performance. 
Table~\ref{table_anno} displays the results compared with training from scratch. The model performance drops sharply when decreasing the training data due to the overfitting issue. Not surprisingly, fine-tuning on top of our pre-trained model can significantly improve performance on all data sizes. The pre-trained model leads to a reduction in the annotation cost by at least $50\%$ on the target tasks compared with training from scratch. The results suggest that our SAR has the potential to improve the performance of 3D tumor segmentation with limited annotations by providing a transferable initialization. 

\section{Conclusion}
In this paper, we propose Scale-Aware Restoration (SAR) Learning, a SSL method based on the prior-knowledge for 3D tumor segmentation. 
Scale discrimination combined with self-restoration is developed to predict the scale information of the input sub-volumes for multi-scale visual patterns learning. We also design an adversarial learning module to learn modality invariant representations. 
The proposed SAR successfully surpasses several 3D SSL methods on two tumor segmentation tasks, especially for small-scale tumor segmentation.
Besides, we demonstrate its effectiveness in terms of data efficiency, performance, and convergence speed. 

\subsubsection{Acknowledgement.}
This work is supported partially by SHEITC (No. 2018-RGZN-02046), 111 plan (No. BP0719010), and STCSM (No. 18DZ2270700).


\bibliographystyle{splncs04}
\bibliography{egbib}

\begin{thebibliography}{10}
\providecommand{\url}[1]{\texttt{#1}}
\providecommand{\urlprefix}{URL }
\providecommand{\doi}[1]{https://doi.org/#1}

\bibitem{Bakas2018}
Bakas, S., et~al.: Identifying the best machine learning algorithms for brain
  tumor segmentation, progression assessment, and overall survival prediction
  in the brats challenge. arXiv preprint arXiv:1811.02629  (2018)

\bibitem{bilic2019liver}
Bilic, P., et~al.: The liver tumor segmentation benchmark (lits). arXiv
  preprint arXiv:1901.04056  (2019)

\bibitem{Chen2019SelfSupervisedLF}
{Chen}, L., {Bentley}, P., {Mori}, K., {Misawa}, K., {Fujiwara}, M.,
  {Rueckert}, D.: Self-supervised learning for medical image analysis using
  image context restoration. Medical Image Analysis  \textbf{58},  101539
  (2019)

\bibitem{DoerschGE15}
Doersch, C., Gupta, A., Efros, A.A.: Unsupervised visual representation
  learning by context prediction. In: IEEE International Conference on Computer
  Vision. pp. 1422--1430 (2015)

\bibitem{Parts2Whole}
Feng, R., et~al.: Parts2whole: Self-supervised contrastive learning via
  reconstruction. In: Domain Adaptation and Representation Transfer, and
  Distributed and Collaborative Learning. pp. 85--95 (2020)

\bibitem{Spyros07728}
Gidaris, S., Singh, P., Komodakis, N.: Unsupervised representation learning by
  predicting image rotations. arXiv preprint arXiv: 1803.07728  (2018)

\bibitem{Haghighi2020LearningSR}
Haghighi, F., Taher, M.R.H., Zhou, Z., Gotway, M.B., Liang, J.: Learning
  semantics-enriched representation via self-discovery, self-classification,
  and self-restoration. In: Medical Image Computing and Computer Assisted
  Intervention. pp. 137--147 (2020)

\bibitem{isensee2019automated}
Isensee, F., J{\"a}ger, P.F., Kohl, S.A., Petersen, J., Maier-Hein, K.H.:
  Automated design of deep learning methods for biomedical image segmentation.
  arXiv preprint arXiv:1904.08128  (2019)

\bibitem{Kingma2014}
Kingma, D., Ba, J.: Adam: A method for stochastic optimization. In:
  International Conference on Learning Representations. vol.~42 (2014)

\bibitem{Litjens2017ASO}
Litjens, G.J.S., et~al.: A survey on deep learning in medical image analysis.
  Medical Image Analysis  \textbf{42},  60--88 (2017)

\bibitem{Setio2015AutomaticDO}
Setio, A.A.A., Jacobs, C., Gelderblom, J., van Ginneken, B.: Automatic
  detection of large pulmonary solid nodules in thoracic ct images. Medical
  physics  \textbf{42},  5642--5653 (2015)

\bibitem{Shin2016}
Shin, H.C., et~al.: Deep convolutional neural networks for computer-aided
  detection: Cnn architectures, dataset characteristics and transfer learning.
  IEEE Transactions on Medical Imaging  \textbf{35}(5),  1285--1298 (2016)

\bibitem{simpson2019large}
Simpson, A.L., et~al.: A large annotated medical image dataset for the
  development and evaluation of segmentation algorithms. arXiv preprint
  arXiv:1902.09063  (2019)

\bibitem{Taleb2020}
Taleb, A., et~al.: 3d self-supervised methods for medical imaging. arXiv
  preprint arXiv:2006.03829  (2020)

\bibitem{Zhou2020ModelsG}
Zhou, Z., Sodha, V., Pang, J., Gotway, M.B., Liang, J.: Models genesis. Medical
  Image Analysis  \textbf{67},  101840 (2021)

\bibitem{Zhuang2019SelfSupervisedFL}
{Zhuang}, X., {Li}, Y., {Hu}, Y., {Ma}, K., {Yang}, Y., {Zheng}, Y.:
  Self-supervised feature learning for 3d medical images by playing a rubik's
  cube. In: Medical Image Computing and Computer Assisted Intervention. pp.
  420--428 (2019)

\bibitem{Zongwei2019}
{Zongwei}, Z., et~al.: Models genesis: Generic autodidactic models for 3d
  medical image analysis. In: Medical Image Computing and Computer Assisted
  Intervention. pp. 384--393 (2019)

\bibitem{iek20163DUL}
Çiçek, {\"O}., Abdulkadir, A., Lienkamp, S.S., Brox, T., Ronneberger, O.: 3d
  u-net: Learning dense volumetric segmentation from sparse annotation. In:
  Medical Image Computing and Computer Assisted Intervention. pp. 424--432
  (2016)

\end{thebibliography}
\end{document}